\documentclass[letterpaper]{article} % DO NOT CHANGE THIS
\usepackage[]{aaai23}  % DO NOT CHANGE THIS
\usepackage{times}  % DO NOT CHANGE THIS
\usepackage{helvet}  % DO NOT CHANGE THIS
\usepackage{courier}  % DO NOT CHANGE THIS
\usepackage[hyphens]{url}  % DO NOT CHANGE THIS
\usepackage{graphicx} % DO NOT CHANGE THIS
\usepackage{natbib}  % DO NOT CHANGE THIS AND DO NOT ADD ANY OPTIONS TO IT
\usepackage{caption} % DO NOT CHANGE THIS AND DO NOT ADD ANY OPTIONS TO IT
\usepackage{algorithm}
\usepackage{algorithmic}
\usepackage{subfigure}
\usepackage{newfloat}
\usepackage{listings}
\DeclareCaptionStyle{ruled}{labelfont=normalfont,labelsep=colon,strut=off} % DO NOT CHANGE THIS
\floatstyle{ruled}
\newfloat{listing}{tb}{lst}{}
\floatname{listing}{Listing}
\title{ConceptX: A Framework for Latent Concept Analysis}

\author{Firoj Alam$^\diamond$ ~ Fahim Dalvi$^\diamond$ ~ Nadir Durrani$^\diamond$ \\
\vspace{3mm} \textbf{Hassan Sajjad$^\clubsuit$\thanks{This work was carried out while the author was at QCRI.} ~ Abdul Rafae Khan$^\spadesuit$ ~ Jia Xu$^\spadesuit$} \\
{\tt \{fialam,faimaduddin,ndurrani\}@hbku.edu.qa} \\ \vspace{3mm}
\textsuperscript{$\diamond$}Qatar Computing Research Institute, HBKU Research Complex, Qatar \\
\textsuperscript{$\clubsuit$}Faculty of Computer Science, Dalhousie University, Canada \\
\textsuperscript{$\spadesuit$}School of Engineering and Science, Steven Institute of Technology, USA \\ 
}

\begin{document}
\maketitle
\begin{abstract}

The opacity of deep neural networks remains a challenge in %the AI community, especially in 
deploying solutions where explanation is as important as precision. %Last few years have seen great strides along the direction of interpreting deep language models. Such tools are based on interactive visualizations \cite{tenney-etal-2020-language,alammar2021ecco}, neuron analysis \cite{neurox-aaai19:demo}, instance-level interpretation \cite{wallace-etal-2019-allennlp}. 
We present \textbf{ConceptX}, a human-in-the-loop framework for interpreting and annotating latent representational space in pre-trained Language Models (pLMs). We use an unsupervised method to discover concepts learned in these models and enable a graphical interface for humans to generate explanations for the concepts. To facilitate the process, we provide auto-annotations of the concepts (based on traditional linguistic ontologies). Such annotations enable development of a linguistic resource that directly represents latent concepts learned within deep NLP models. These include not just traditional linguistic concepts, but also task-specific or sensitive concepts (words grouped based on gender or religious connotation) that helps the annotators to mark bias in the model. The framework consists of two parts (i) concept discovery\footnote{Code available at: \url{https://github.com/hsajjad/ConceptX}} and (ii) annotation platform.\footnote{Main platform: \url{https://micromappers.qcri.org/}, where project can be created for a new task. Here is an example of the BERT concept annotation task: \url{https://micromappers.qcri.org/project/nx-concept-annotation}}\textsuperscript{,}\footnote{\url{https://youtu.be/t8UCP6WPoYg}}

\end{abstract}

\section{Introduction}
\label{sec:introduction}
% Motivation of latent clusters, model centric interpretation, benefit of human in the loop to interpret models

% \begin{itemize}
    % \item Goal 1: Deep NLP, black-box, latent concept as a new framework ICLR cite, NAACL cite. Interpretation is dependent on pre-defined linguistic concepts, human-bias, analyze latent concepts by having human in the loop
    % \item Framework setup and pieces: clustering, auto-label, trivial concepts, 
    % \item Explain clustering,
    % \item explain auto-labeling A lot of work has been done to use pre-defined concepts that we can use to facilitate the humans
    % \item manual annotations, questions, sibling clusters
    % \item product: BCN corpus, potential applications, human in the loop deployment of AI systems to highlight potential bias 
% \end{itemize}

% Large-scale deep neural network (DNN) models have achieved state-of-the-art performance in downstream natural language processing (NLP) tasks. However, their black-box nature remains a challenge for their wide-scale adaptation. %in deploying AI solutions where fairness and trust are as important as the model's precision. 
% The revolution of DNN models has subsequently spurred a plethora of work in interpreting these models. To this end, a lot of work has been carried towards post-hoc representation analysis to uncover linguistic phenomena that are captured as DNNs are trained towards any NLP task \cite{adi2016fine,belinkov:2017:acl,conneau2018you,liu-etal-2019-linguistic,tenney-etal-2019-bert}. 

Work done on representation analysis undercover linguistic phenomena that are captured as DNNs are trained towards any NLP task \cite{belinkov:2017:acl,liu-etal-2019-linguistic, tenney-etal-2019-bert, durrani-etal-2020-analyzing}. A downside of previous methodologies 
%A downside of the work carried out in representation analysis
is that their scope is limited to pre-defined concepts that only reinforce traditional linguistic knowledge and do not reflect on the novel concepts learned by the model, therefore resulting in a narrow view of the knowledge captured by pre-trained Language Models. %Another weakness of using pre-defined concepts is the involvement of human bias in selecting the concept which may result in a misleading interpretation. 

We do away with this problem by presenting \textbf{a Framework for Latent Concept Analysis (LCA) in deep NLP Models.} %Our framework enables human-in-loop interpretation 
% We cluster contextualized representations in high-dimensional space using hierarchical clustering to identify groups of related words. Our framework enables a platform to carry manual annotation of the discovered concepts \cite{dalvi2022discovering}. To facilitate the annotation process, we also auto-align the latent concepts with the pre-defined concepts using our \texttt{ConceptX} alignment framework \cite{sajjad:naacl:2022}. In the following, we briefly describe each piece of our setup.
Our framework is composed of two modules: i) Concept Discovery and ii) Annotation Platform. We discover latent concepts in deep NLP models using an unsupervised approach \cite{dalvi2022discovering} and enable a human-in-the-loop platform to annotate these concepts. The framework is facilitated by auto-labeling the concepts by aligning them to the linguistic ontologies \cite{sajjad:naacl:2022}.

\section{Platform Overview}
\label{sec:disinfo_system}

%\subsection{Pipeline}

%In Figure \ref{fig:pipeline}, we present the pipeline of the proposed framework, which takes input text and model as input, extract representations for each token from the input text. Then, the extracted representations are used for clustering. The extracted clusters are then visually inspected for manual verification and labeling. 

In Figure \ref{fig:pipeline} we present the pipeline of our framework and below we describe these modules:

%It is composed of two parts: i) concept discovery, ii) annotation platform.

\begin{figure*}[h]
\centering
\includegraphics[width=0.7\textwidth]{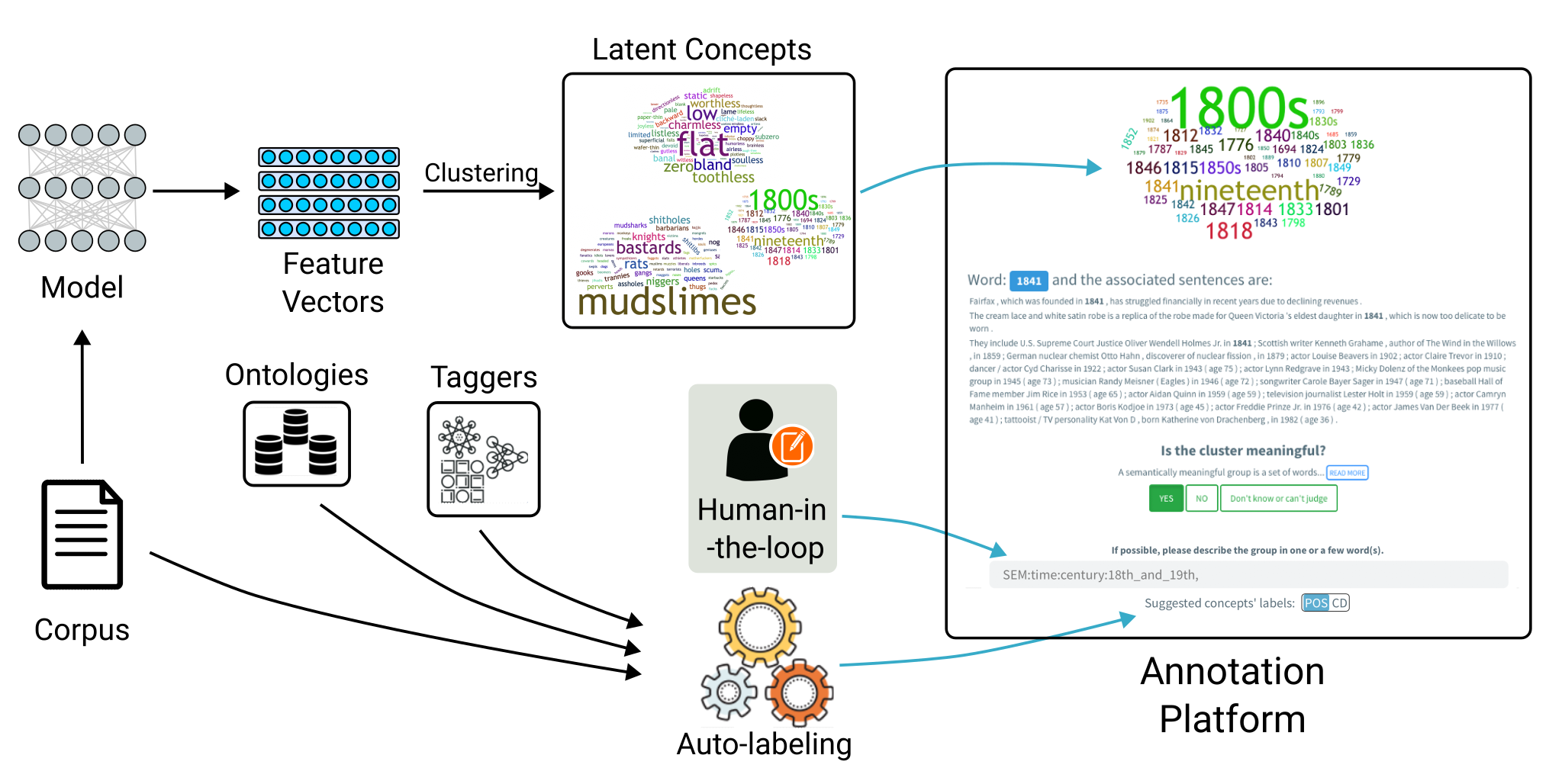}
\vspace{-3mm}
\caption{Complete pipeline of the proposed framework for concept analysis in the trained models.}
\label{fig:pipeline}
\end{figure*}

\subsection{Concept Discovery}

%\subsubsection{Encoded Representation}
%\label{sssec:encoded_representation}

Our method is based on an unsupervised approach to discover latent concepts in the representational space of the pre-trained models \cite{dalvi2022discovering}. We generate feature vectors (contextualized representation) %from the model 
by doing a forward-pass on the model, and cluster representations using agglomerative hierarchical clustering~\citep{gowda1978agglomerative} to discover the encoded concepts. Our hypothesis is that contextualized word representations learned within pLMs capture \emph{meaningful} groupings based on a coherent concept such as lexical, syntactic and semantic similarity, or any task or data specific pattern that groups the words together \cite{neuronSurvey}. %See Figure \ref{fig:sample_polarity_clusters} for two example concepts from the models trained for sentiment analysis and hate-speech detection tasks. 

\subsection{Annotation Platform}
\label{sssec:auto_labels}
% \todo[inline]{Are we talking about auto labels in this paper? Or there could be a separate paper?}

Once we have discovered the latent concepts captured in the model, we provide a human-in-the-loop framework to analyze %enable annotations of 
and annotate these concepts. To facilitate the effort, we auto-label the concepts with their linguistic connotations wherever applicable.

% we want to analyze and annotate them by having human-in-the-loop. To facilitate the effort, we auto-annotate the concepts with their linguistic connotations wherever we can. In the following, we describe these two pieces of our pipeline.

\subsubsection{Annotation Guidelines} 

%We prepared detailed instructions to guide the annotators. 
Our annotation consists of two questions, \textbf{Q1:} Does the cluster represent a meaningful concept? \textbf{Q2:} Can the neighboring clusters be combined to form a meaningful concept? We provide annotators with specific instructions with examples of what constitutes a meaningful concept based on our definition \cite{neuronSurvey}. The word cluster is represented in a form of a word cloud of examples -- See Figure \ref{fig:sample_polarity_clusters}), where the relative size of a word in the word cloud depends on the frequency of the word in the data. To understand the context of each word in the cluster we also facilitate the annotators with associated sentences %from 
in the dataset. More context is enabled through Google search within our framework, where the the annotator can look-up for meaning and more contextual information on the words in the concept. Due to hierarchical clustering, we often expect neighboring clusters to capture very similar concepts that can be conjoined. For example neighboring clusters capturing Hindu and Muslim names can be combined together to form a Hindu-Muslim names concept. Our goal in Q2 is to capture such concepts. Having such an annotation facilitates analysis at a hierarchical level.

\subsubsection{Auto-labeling}

Pre-trained language models have %been 
shown to learn rich linguistic concepts \cite{belinkov:2017:acl,liu-etal-2019-linguistic,tenney-etal-2019-bert, dalvi-2020-CCFS}. %such as morphology, syntax, and semantics. Much of these linguistic phenomena are carried forward even as the models are fine-tuned towards downstream tasks \cite{merchant-etal-2020-happens,mosbach-etal-2020-interplay,durrani-etal-2021-transfer}. 
To prevent the human effort of re-annotating such concepts, we integrate an alignment framework \cite{sajjad:naacl:2022} in our platform. We annotated the training data used to obtain latent concepts, with core linguistic concepts (e.g., parts-of-speech, word-net tags etc.) and map the latent concepts to linguistic concepts through an alignment function.

\section{Use Cases} 
\label{sec:application}

% We used our platform to annotate concepts from BERT-base-cased. 

\subsection{BCN Development}

The BERT conceptNet (BCN) dataset\footnote{\url{https://neurox.qcri.org/projects/bert-concept-net.html}} development is an example of the utilization of the framework. We had a group of 6 linguists annotate encoded concepts in the BERT-base-cased model. The resulting dataset: BCN is a unique multi-faceted resource consisting of 174 concept labels with a total of 997,195 annotated instances. It can be used by the community to benchmark efforts on interpretability using model's concept instead of relying on extrinsic concepts. The hierarchy present in the concept labels provides flexibility to use data at different granularities.

\begin{figure}[t]
    \begin{subfigure}
    \centering
    \includegraphics[width=.49\linewidth]{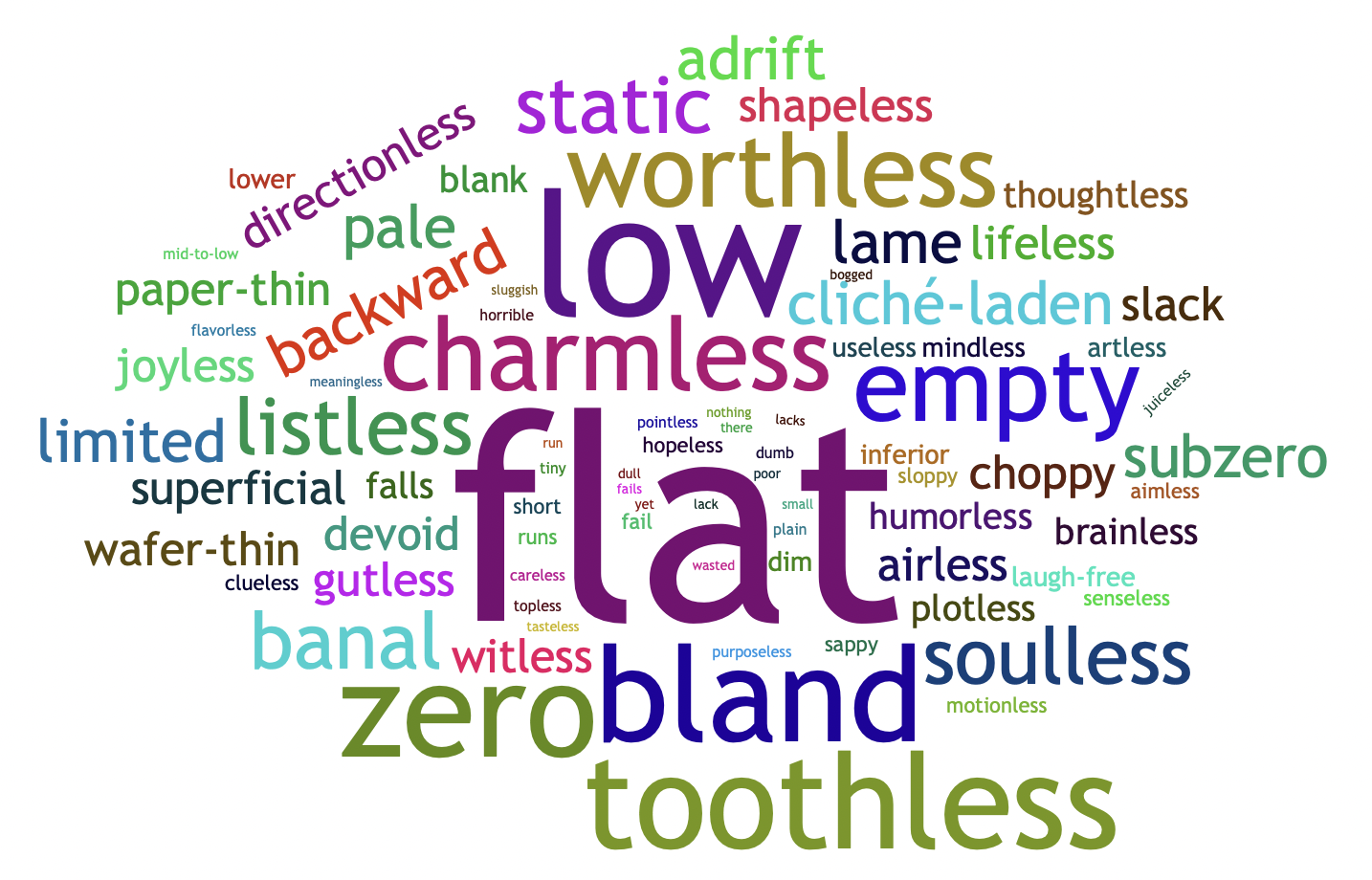}
    \label{fig:xlm-positive}
    %\caption{XLM-SST Layer 12, c15}
    \end{subfigure}
    \begin{subfigure}
    \centering
    \includegraphics[width=.49\linewidth]{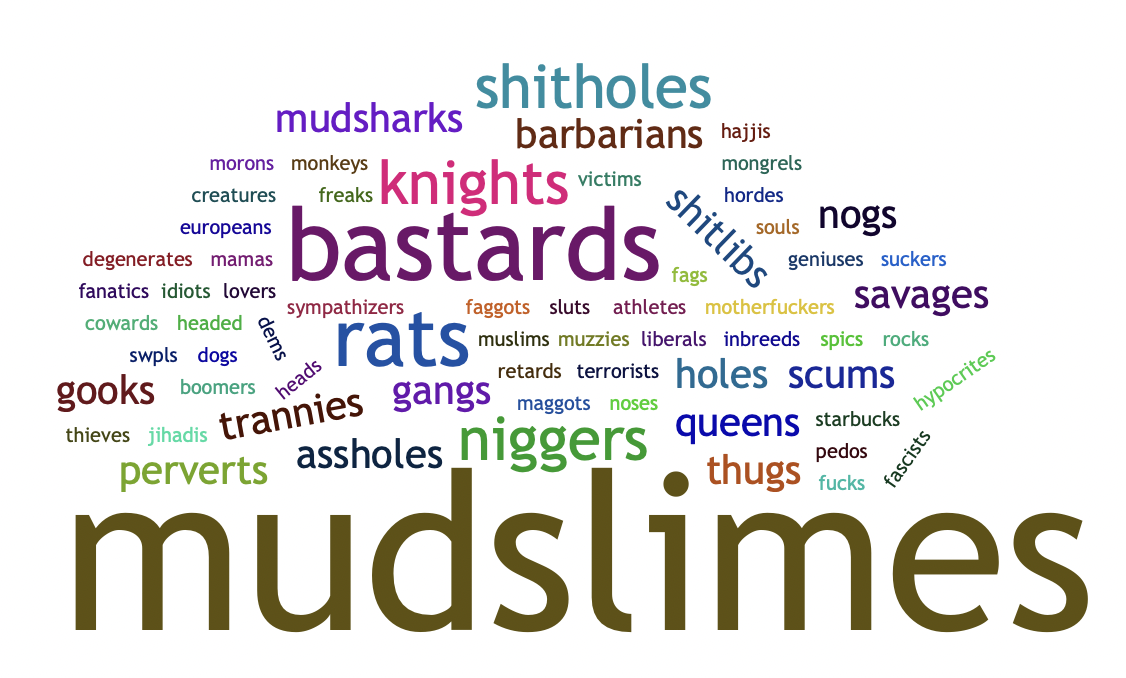}
    \label{fig:xlm-negative}
    %\caption{XLM-SST Layer 10, c16}
    \end{subfigure}
    \vspace{-4mm}
    %\caption{Positive (a) and Negative (b) Polarity Concepts}
    \caption{Polarity Concepts: Negative Sentiment  (SST-2), Toxic (right) (HSD)}
    \label{fig:sample_polarity_clusters}
    \vspace{-2mm}
\end{figure}

\subsection{Task Specific Model Analysis}
In addition to the BCN development, we analyzed %popular transformer architecture 
two transformers -- BERT-base-cased \cite{devlin-etal-2019-bert} and  XLM-RoBERT \cite{xlm-roberta} models, fine-tuned towards
two %downstream 
tasks: sentiment analysis \cite[SST-2,][]{socher-etal-2013-recursive} and the hate speech detection~\cite[HSD,][]{mathew2021hatexplain}. Our framework helps visualize how the models segregates negative and positive polarity concepts, in the final layers of the models. See Figure \ref{fig:sample_polarity_clusters} for examples of polarity concepts in the two tasks and \cite{durrani-EMNLP-22} for details.

\section{Related Work}
\label{sec:related_work}

A number of toolkits have been made available to carry  analysis of neural network models. Google’s What-If tool \cite{Wexler_2019} inspects machine learning models
and provides users an insight of the trained model based
on the predictions. Seq2Seq-Vis
\cite{strobelt-etal-2018-debugging} enables the user to trace back the prediction decisions to the input in NMT models. Captum \cite{Captum} provides generic implementations of a number of gradient and perturbation-based attribution algorithms. NeuroX \cite{neurox-aaai19:demo} and Ecco~\cite{alammar2021ecco} use probing classifiers to examine the representations in pLMs. %pre-trained language models. 
\citet{tenney-etal-2020-language} facilitates debugging of pLMs through intractive visualizations. Our goal is slightly different from these toolkits. Along with the analysis of the latent spaces within pLMs, we intend to annotate latent ontologies learned by these models using human-in-the-loop facilitated by traditional linguistic knowledge. We believe such annotations could be useful to benchmark efforts in interpretation and provide a more realistic view of what these models are capturing.

\bibliography{bib/anthology,bib/custom}

\begin{thebibliography}{20}
\providecommand{\natexlab}[1]{#1}

\bibitem[{Alammar(2021)}]{alammar2021ecco}
Alammar, J. 2021.
\newblock Ecco: an open source library for the explainability of transformer
  language models.
\newblock In \emph{Proceedings of the 59th Annual Meeting of the Association
  for Computational Linguistics and the 11th International Joint Conference on
  Natural Language Processing: System Demonstrations}, 249--257.

\bibitem[{Belinkov et~al.(2017)Belinkov, Durrani, Dalvi, Sajjad, and
  Glass}]{belinkov:2017:acl}
Belinkov, Y.; Durrani, N.; Dalvi, F.; Sajjad, H.; and Glass, J. 2017.
\newblock What do Neural Machine Translation Models Learn about Morphology?
\newblock In \emph{Proceedings of the 55th Annual Meeting of the Association
  for Computational Linguistics}, ACL~'17, 861--872. Vancouver, Canada:
  Association for Computational Linguistics.

\bibitem[{Conneau et~al.(2020)Conneau, Khandelwal, Goyal, Chaudhary, Wenzek,
  Guzm{\'a}n, Grave, Ott, Zettlemoyer, and Stoyanov}]{xlm-roberta}
Conneau, A.; Khandelwal, K.; Goyal, N.; Chaudhary, V.; Wenzek, G.; Guzm{\'a}n,
  F.; Grave, E.; Ott, M.; Zettlemoyer, L.; and Stoyanov, V. 2020.
\newblock Unsupervised Cross-lingual Representation Learning at Scale.
\newblock In \emph{Proceedings of the 58th Annual Meeting of the Association
  for Computational Linguistics}, 8440--8451. Association for Computational
  Linguistics.

\bibitem[{Dalvi et~al.(2022)Dalvi, Khan, Alam, Durrani, Xu, and
  Sajjad}]{dalvi2022discovering}
Dalvi, F.; Khan, A.~R.; Alam, F.; Durrani, N.; Xu, J.; and Sajjad, H. 2022.
\newblock Discovering Latent Concepts Learned in {BERT}.
\newblock In \emph{Proceedings of the Tenth International Conference on
  Learning Representations}, ICLR~'22. Online.

\bibitem[{Dalvi et~al.(2019)Dalvi, Nortonsmith, Bau, Belinkov, Sajjad, Durrani,
  and Glass}]{neurox-aaai19:demo}
Dalvi, F.; Nortonsmith, A.; Bau, D.~A.; Belinkov, Y.; Sajjad, H.; Durrani, N.;
  and Glass, J. 2019.
\newblock NeuroX: A Toolkit for Analyzing Individual Neurons in Neural
  Networks.
\newblock In \emph{Proceedings of the AAAI Conference on Artificial
  Intelligence}, AAAI~'19, 9851--9852. Honolulu, USA.

\bibitem[{Dalvi et~al.(2020)Dalvi, Sajjad, Durrani, and
  Belinkov}]{dalvi-2020-CCFS}
Dalvi, F.; Sajjad, H.; Durrani, N.; and Belinkov, Y. 2020.
\newblock Analyzing Redundancy in Pretrained Transformer Models.
\newblock In \emph{Proceedings of the 2020 Conference on Empirical Methods in
  Natural Language Processing (EMNLP-2020)}. Online.

\bibitem[{Devlin et~al.(2019)Devlin, Chang, Lee, and
  Toutanova}]{devlin-etal-2019-bert}
Devlin, J.; Chang, M.-W.; Lee, K.; and Toutanova, K. 2019.
\newblock {BERT}: Pre-training of Deep Bidirectional Transformers for Language
  Understanding.
\newblock In \emph{Proceedings of the 2019 Conference of the North {A}merican
  Chapter of the Association for Computational Linguistics: Human Language
  Technologies}, NAACL-HLT~'19, 4171--4186. Minneapolis, Minnesota, USA:
  Association for Computational Linguistics.

\bibitem[{Durrani et~al.(2022)Durrani, Sajjad, Dalvi, and
  Alam}]{durrani-EMNLP-22}
Durrani, N.; Sajjad, H.; Dalvi, F.; and Alam, F. 2022.
\newblock On the Transformation of Latent Space in Fine-Tuned NLP Models.
\newblock In \emph{Proceedings of the 2022 Conference on Empirical Methods in
  Natural Language Processing}, EMNLP. Abu Dhabi, UAE: Association for
  Computational Linguistics.

\bibitem[{Durrani et~al.(2020)Durrani, Sajjad, Dalvi, and
  Belinkov}]{durrani-etal-2020-analyzing}
Durrani, N.; Sajjad, H.; Dalvi, F.; and Belinkov, Y. 2020.
\newblock Analyzing Individual Neurons in Pre-trained Language Models.
\newblock In \emph{Proceedings of the 2020 Conference on Empirical Methods in
  Natural Language Processing}, EMNLP, 4865--4880. Online: Association for
  Computational Linguistics.

\bibitem[{Gowda and Krishna(1978)}]{gowda1978agglomerative}
Gowda, K.~C.; and Krishna, G. 1978.
\newblock Agglomerative clustering using the concept of mutual nearest
  neighbourhood.
\newblock \emph{Pattern recognition}, 10(2): 105--112.

\bibitem[{Kokhlikyan et~al.(2020)Kokhlikyan, Miglani, Martin, Wang, Alsallakh,
  Reynolds, Melnikov, Kliushkina, Araya, Yan, and Reblitz-Richardson}]{Captum}
Kokhlikyan, N.; Miglani, V.; Martin, M.; Wang, E.; Alsallakh, B.; Reynolds, J.;
  Melnikov, A.; Kliushkina, N.; Araya, C.; Yan, S.; and Reblitz-Richardson, O.
  2020.
\newblock Captum: A unified and generic model interpretability library for
  PyTorch.

\bibitem[{Liu et~al.(2019)Liu, Gardner, Belinkov, Peters, and
  Smith}]{liu-etal-2019-linguistic}
Liu, N.~F.; Gardner, M.; Belinkov, Y.; Peters, M.~E.; and Smith, N.~A. 2019.
\newblock Linguistic Knowledge and Transferability of Contextual
  Representations.
\newblock In \emph{Proceedings of the 2019 Conference of the North {A}merican
  Chapter of the Association for Computational Linguistics: Human Language
  Technologies}, NAACL~'19, 1073--1094. Minneapolis, Minnesota, USA:
  Association for Computational Linguistics.

\bibitem[{Mathew et~al.(2021)Mathew, Saha, Yimam, Biemann, Goyal, and
  Mukherjee}]{mathew2021hatexplain}
Mathew, B.; Saha, P.; Yimam, S.~M.; Biemann, C.; Goyal, P.; and Mukherjee, A.
  2021.
\newblock Hatexplain: A benchmark dataset for explainable hate speech
  detection.
\newblock In \emph{Proceedings of the AAAI Conference on Artificial
  Intelligence}, volume~35, 14867--14875.

\bibitem[{Sajjad, Durrani, and Dalvi(2021)}]{neuronSurvey}
Sajjad, H.; Durrani, N.; and Dalvi, F. 2021.
\newblock {Neuron-level Interpretation of Deep NLP Models: A Survey}.
\newblock \emph{CoRR}, abs/2108.13138.

\bibitem[{Sajjad et~al.(2022)Sajjad, Durrani, Dalvi, Alam, Khan, and
  Xu}]{sajjad:naacl:2022}
Sajjad, H.; Durrani, N.; Dalvi, F.; Alam, F.; Khan, A.~R.; and Xu, J. 2022.
\newblock Analyzing Encoded Concepts in Transformer Language Models.
\newblock In \emph{Proceedings of the 2022 Conference of the North American
  Chapter of the Association for Computational Linguistics}, NAACL~'22.
  Seattle, Washington, USA: Association for Computational Linguistics.

\bibitem[{Socher et~al.(2013)Socher, Perelygin, Wu, Chuang, Manning, Ng, and
  Potts}]{socher-etal-2013-recursive}
Socher, R.; Perelygin, A.; Wu, J.; Chuang, J.; Manning, C.~D.; Ng, A.; and
  Potts, C. 2013.
\newblock Recursive Deep Models for Semantic Compositionality Over a Sentiment
  Treebank.
\newblock In \emph{Proceedings of the 2013 Conference on Empirical Methods in
  Natural Language Processing}, 1631--1642. Seattle, Washington, USA:
  Association for Computational Linguistics.

\bibitem[{Strobelt et~al.(2018)Strobelt, Gehrmann, Behrisch, Perer, Pfister,
  and Rush}]{strobelt-etal-2018-debugging}
Strobelt, H.; Gehrmann, S.; Behrisch, M.; Perer, A.; Pfister, H.; and Rush, A.
  2018.
\newblock Debugging Sequence-to-Sequence Models with {S}eq2{S}eq-Vis.
\newblock In \emph{Proceedings of the 2018 {EMNLP} Workshop {B}lackbox{NLP}:
  Analyzing and Interpreting Neural Networks for {NLP}}, 368--370. Brussels,
  Belgium: Association for Computational Linguistics.

\bibitem[{Tenney, Das, and Pavlick(2019)}]{tenney-etal-2019-bert}
Tenney, I.; Das, D.; and Pavlick, E. 2019.
\newblock {BERT} Rediscovers the Classical {NLP} Pipeline.
\newblock In \emph{Proceedings of the 57th Annual Meeting of the Association
  for Computational Linguistics}, 4593--4601. Florence, Italy: Association for
  Computational Linguistics.

\bibitem[{Tenney et~al.(2020)Tenney, Wexler, Bastings, Bolukbasi, Coenen,
  Gehrmann, Jiang, Pushkarna, Radebaugh, Reif, and
  Yuan}]{tenney-etal-2020-language}
Tenney, I.; Wexler, J.; Bastings, J.; Bolukbasi, T.; Coenen, A.; Gehrmann, S.;
  Jiang, E.; Pushkarna, M.; Radebaugh, C.; Reif, E.; and Yuan, A. 2020.
\newblock The Language Interpretability Tool: Extensible, Interactive
  Visualizations and Analysis for {NLP} Models.
\newblock In \emph{Proceedings of the 2020 Conference on Empirical Methods in
  Natural Language Processing: System Demonstrations}, 107--118. Online:
  Association for Computational Linguistics.

\bibitem[{Wexler et~al.(2019)Wexler, Pushkarna, Bolukbasi, Wattenberg, Viegas,
  and Wilson}]{Wexler_2019}
Wexler, J.; Pushkarna, M.; Bolukbasi, T.; Wattenberg, M.; Viegas, F.; and
  Wilson, J. 2019.
\newblock The What-If Tool: Interactive Probing of Machine Learning Models.
\newblock \emph{{IEEE} Transactions on Visualization and Computer Graphics},
  1--1.

\end{thebibliography}
% \bibliographystyle{aaai23}

% \newpage
% \clearpage
% \section*{Appendix} 
% \label{sec:appendix}
% \appendix
% \input{supplemental_material}

\end{document}